\documentclass{article}

     \PassOptionsToPackage{numbers, compress}{natbib}

     \usepackage[final]{neurips_2022}

\usepackage[utf8]{inputenc} 
\usepackage[T1]{fontenc}    
\usepackage{hyperref}       
\usepackage{url}            
\usepackage{booktabs}       
\usepackage{amsfonts}       
\usepackage{nicefrac}       
\usepackage{microtype}      
\usepackage{xcolor}         
\usepackage{latexsym} 
\usepackage{multirow}
\usepackage{graphicx}
\usepackage{wrapfig}
\usepackage{amsmath}
\usepackage{mathtools}
\usepackage{xspace}
\usepackage{floatflt}
\usepackage{tikz}

    \title{A Causal Analysis of Harm}
\author{
 Sander Beckers\\
 Cluster of Excellence in Machine Learning\\
 University of T\"ubingen and\\
 Munich Center for Mathematical Philosophy, LMU\\
 srekcebrednas@gmail.com
 \And
  Hana Chockler\\
  causaLens and\\
  Department of Informatics\\
  King's College London\\
  hana.chockler@kcl.ac.uk\\
  \And
  Joseph Y. Halpern\\
  Computer Science Department\\
  Cornell University\\
  halpern@cs.cornell.edu
  }

\newcommand{\Scal}{{\cal S}}

\newcommand{\bd}{\begin{definition}}
\newcommand{\ed}{\end{definition}}
\newcommand{\be}{\begin{enumerate}}
\newcommand{\bi}{\begin{itemize}}
\newcommand{\ee}{\end{enumerate}}
\newcommand{\ei}{\end{itemize}}

\newcommand{\U}{{\cal U}}

\newcommand{\V}{{\cal V}}
\newcommand{\R}{{\cal R}}

\newcommand{\stam}[1]{}

\newcommand{\commentout}[1]{}

\newtheorem{definition}{Definition}

\newtheorem{theorem}{Theorem}[section]
\newtheorem{lemma}{Lemma}
\newtheorem{proposition}[theorem]{Proposition}
\newtheorem{example}{Example}

\newtheorem{claim}{Claim}

\newcommand{\wbox}{\mbox{$\sqcap$\llap{$\sqcup$}}}

\newcommand{\clm}{\begin{claim}}
\newcommand{\eclm}{\end{claim}}

\newcommand{\union}{\cup}

\newcommand{\eg}{e.g.,~}

\newcommand{\thm}{\begin{theorem}}
\newcommand{\pro}{\begin{proposition}}
\newcommand{\ethm}{\end{theorem}}

\definecolor{darkred}{rgb}{0.65,0,0}

\newcommand{\F}{{\cal F}}

\newcommand{\sat}{\models}

\newcommand{\lem}{\begin{lemma}}
\newcommand{\elem}{\end{lemma}}
\newcommand{\epro}{\end{proposition}}

\newcommand{\dfn}{\begin{definition}}
\newcommand{\edfn}{\end{definition}}

\newcommand{\xam}{\begin{example}}
\newcommand{\exam}{\end{example}}
\renewcommand{\citeyear}{\cite}

\begin{document}

\maketitle

\begin{abstract}
As autonomous systems rapidly become ubiquitous,
there is a growing need for a legal and regulatory framework 
that addresses when and how such a system harms someone. 
There have been several attempts within the philosophy literature to define harm, but none of them has proven capable of dealing with
the many examples that have been presented, 
leading some to suggest that the notion of harm should be
abandoned and ``replaced by more well-behaved notions''. As harm is
generally something that is caused, most of these definitions have
involved causality at some level. Yet surprisingly, none of them makes
use of causal models and the definitions of actual causality that they
can express. In this paper we formally define a qualitative notion of harm that
uses causal models and is based on a well-known definition of actual
causality \cite{Hal48}.
The key features of our definition 
are that it is based on \emph{contrastive} causation
and uses a default utility to which the utility of actual outcomes
is compared. We show that our definition is able to handle
the examples from the literature, and illustrate its importance
for reasoning about situations involving autonomous systems. 
\end{abstract}

\section{Introduction}\label{sec:intro}
\vspace{-0.1cm}

The notion that one should not cause harm is a central tenet in many religions;
it is enshrined in the medical profession's Hippocratic Oath, which
states explicitly ``I will do no harm or injustice to
[my
  patients]''~\cite{Hipp02}
it is also a critical element in the law.
Not surprisingly, there have been many attempts in the philosophy
literature to define harm.  
Motivated by the observation that we speak of ``causing harm'', most of these have involved causality at some level.
All these attempts have encountered difficulties.  
Indeed, Bradley \citeyear{Bra12} says:

\textit{Unfortunately, when we look at
attempts to explain the nature of harm, we find a mess.  The most
widely discussed account, the comparative account, faces
counterexamples that seem fatal.  But no alternative account has
gained any currency.  My diagnosis is that the notion of harm 
is a Frankensteinian jumble \ldots It should be replaced by other more
well-behaved notions.    
}

The situation has not improved much since Bradley's paper
(see, e.g., recent accounts like \cite{CJR21,Feit19}).
Yet the legal and regulatory aspects of harm are becoming particularly
important now, as
autonomous systems become increasingly more prevalent.
In fact, the new proposal for Europe's AI
act \cite{EuropeAIAct} contains over 25 references to ``harm'' or
``harmful'', saying such things as ``\ldots it is appropriate to
classify [AI systems] as high-risk if, in the light of their intended
purpose, they pose a high risk of harm to the health and safety or the
fundamental rights of persons \ldots'' \cite[Proposal preamble, clause (32)]{EuropeAIAct}. 
Moreover, the European Commission recognized that if harm is to play such a
crucial role, it must be defined carefully, saying 
``Stakeholders also highlighted that \ldots it is important to define
\ldots ‘harm’ \cite[Part 2, Section 3.1]{EuropeAIAct}.
Legislative bodies in the UK are also discussing the question of harm
and who caused harm in 
the
case of
accidents involving autonomous vehicles. 
The Law Commission of England and Wales and the Scottish Law
Commission are recommending 
that
drivers of self-driving cars should not be legally responsible for
crashes; rather, the onus should 
lie with
the manufacturer~\cite{LC22}.
In particular, if there is harm then this is caused by the manufacturers.
The manufacturers translate this recommendation to a standard
according to which
the driver does not even have to pay attention while 
at the wheel. If a complex situation arises on the road requiring
the driver's attention, the car will notify the driver, giving them
$10$ seconds to  
take control. If the driver does not react in time, the car
will flash emergency lights, slow down, and eventually stop~\cite{BBC21}.
Consider the following example (to which we return later).
\xam[{\bf Autonomous Car}]\label{ex:car}
\emph{
An autonomous car detects an unexpected stationary car in front of it
on a highway. It could alert the driver Bob, who would then
have to react within $10$ seconds. However, 
$10$ seconds is too long: the car will crash into the stationary car
within $8$ seconds. The autonomous car's algorithm directs it to crash
into the safety fence on the side of 
the highway, injuring Bob. Bob claims that he was harmed by the
car. Moreover, he also claims that, if alerted, he would have been able
to find a better solution that would not have resulted in his being
injured (e.g., 
swerving into the incoming traffic 
then back to his own lane after passing the stationary car). We assume
that if the 
autonomous car had done nothing and collided 
with the stationary car, both drivers would have been injured much
more severely.
}
\hfill \wbox
\exam
\vspace{-0.2cm}
While the causal model depicting this story is fairly straightforward,
the decision on whether harm was caused to Bob, and if yes, who 
or what caused
the harm, is far less clear.  Indeed, the philosophy literature seems
to suggest that trying to determine this systematically is a lost cause.
But as this example illustrates, the stakes of having a
well-defined notion of harm have become much higher with the advent of
automated decision-making. In contrast to human agents, such systems
do not have an informal understanding of harm that informs their
actions; so we need a formal definition.
Situations like that described in Example~\ref{ex:car} are bound to
arise frequently in the interaction of 
autonomous systems with human users, in a variety of domains. 
We briefly outline two of those.

Imagine a UAV used by the military has to decide whether or not it
should bomb a suspected enemy encampment. The problem is that the
target is not clearly identified, because there are two camps close to
each other: one consisting of civilian refugees, another consisting of
a rebel group that is about to launch a deadly attack on the refugee
camp, killing all of its inhabitants. The UAV's decision is based only
on the expected utility of the refugees, and therefore it bombs the
camp. Tragically, as it turns out, the camp was that of the
refugees. Here we have the intuition that the UAV harmed these
refugees, despite the fact that both actions would have led to all the
refugees being killed.
Examples in which one event (the bombing) preempts another event (the attack) from causing an outcome
 are known as {\em Late Preemption} examples in
the causality literature; we discuss them later in the paper.

In the healthcare domain, autonomous systems are used 
for, among other things, 
classifying MRI brain images suspected of containing a tumor. 
If an image is classified as having 
a tumor, the system 
decides whether to recommend a
surgery. 
While the overall accuracy of the system is superior to
that of humans, in some instances  
the system overlooks an operable tumor.
Imagine a patient who has such a tumor and
dies from brain cancer as the result of not undergoing 
surgery, leading to
a dispute between the patient's family and the hospital
regarding whether
the patient was harmed. Even if both parties
agree that the patient would probably have been alive if the diagnosis
had been performed by a human, the hospital might claim that using the
system is the optimal policy, and therefore one should compare
the actual outcome only to those that could have occurred under the
policy. 

Fortunately, the formal tools at our disposal to develop a formal
notion of harm have also improved over the past few years; we take
full advantage of these  developments in this paper. 
Concretely, we provide a formal definition of harm that we believe
deals with all the concerns 
that have been raised, 
seems
to match our intuitions well, and 
connects closely
 to work on decision theory and utility.
Here we briefly give a 
high-level overview of the key features of our approach and how they
deal with the problems raised in the earlier papers.

There is one set of problems that arise from using counterfactuals
that also arise with causality, and can be dealt with using the by-now
standard approaches in defining causality.  For example, Carlson,
Johansson, and Risberg \citeyear{CJR21} raise a number of problems with
defining harm causally that are solved by simply applying the
definition of actual causality given by Halpern \citeyear{Hal47,Hal48}.
The issue of whether failing to take an action can be viewed as causing
harm (e.g., can failing to water a neighbor's plants after promising
to do so be viewed as causing harm) can also be dealt with by using
the standard definition of causality (which allows lack of an action to
be a cause).%

We remark that Richens, Beard, and Thompson \citeyear{RBT22} (RBT from
now on) 
also
recently observed that using causality appropriately could deal with 
some of the problems raised in the harm literature.%
\footnote{Indeed, a talk by Jonathan Richens that discussed these
issues
(which are also discussed in \cite{RBT22})
was attended by one of the authors of this paper,
and it motivated us to look carefully at harm.}
They also give a formal definition of harm that uses causal models,
but it does not make use of a sophisticated definition of actual
causality such as the one given by Halpern 
\citeyear{Hal47,Hal48}.  (See
Section~\ref{sec:RBT} 
for a comparison of our approach to theirs and more discussion of
this issue.)  RBT focus on the more quantitative,
probabilistic 
aspects
 of harm. 
 \commentout{
 While we believe that a quantitative account is
extremely important (and we are currently working on extending our
account to a more quantitative setting; see 
Section~\ref{sec:con}), 
it is critical to start
with a qualitative account, in part because 
much of the discussion of harm (including the car example above) is
basically qualitative, and because we think that it will be easier to
get a good quantitative account once we have a good qualitative account
to build on.
}
We also believe that a quantitative account is
extremely important; we offer such an account in \cite{BCH22b}.
Conceptually though, the qualitative notion comes first: only after
establishing whether or not there was harm does it make sense to ask
how much harm occurred. 
Indeed, our quantitative account generalizes the qualitative account
we develop here in several ways (see Section~\ref{sec:con}).  

In any case, just applying the definition of causality does not deal with 
all problems.  The other key step that we take is to assume that there
exists a \emph{default} utility.
Roughly speaking, we define an event to cause harm whenever it
causes the utility of the outcome to be lower than the default
utility. 
The default may be context-dependent, and there may be disagreement
about what the default should be. 
 We view that as a feature of our
definition.  For example, we can capture the fact that people
 disagree
about whether a doctor euthanizing a patient in great pain causes harm 
by taking it to be a disagreement about what the appropriate default
should be. 
Likewise, the dispute between the family and the hospital described
above can be modeled as a disagreement about the right default. 
 Moreover, by explicitly bringing utility into the picture,
 we can connect issues that that have been discussed at length
 regarding utility (e.g., what the appropriate discount factor to
 apply to the utility of future generations is) to issues of harm.   
 
\vspace{-0.3cm}
\section{Causal Models and Actual Causality}\label{sec:models}
\vspace{-0.1cm}
We start with a review of causal models \cite{HP01b}, since they play
a critical role in our definition of harm.  The material in this
section is largely taken from \cite{Hal48}.
We assume that the world is described in terms of 
variables and their values.  Some variables may have a causal
influence on others. This influence is modeled by a set of {\em
  structural equations}. It is conceptually useful to split the
variables into two sets: the {\em exogenous\/} variables, whose values
are determined by factors outside the model, and the
{\em endogenous\/} variables, whose values are ultimately determined by
the exogenous variables.  The structural equations describe how these
values are determined.

Formally, a \emph{causal model} $M$
is a pair $(\Scal,\F)$, where $\Scal$ is a \emph{signature}, which explicitly
lists the endogenous and exogenous variables  and characterizes
their possible values, and $\F$ defines a set of \emph{(modifiable)
structural equations}, relating the values of the variables.  
A signature $\Scal$ is a tuple $(\U,\V,\R)$, where $\U$ is a set of
exogenous variables, $\V$ is a set 
of endogenous variables, and $\R$ associates with every variable $Y \in 
\U \union \V$ a nonempty set $\R(Y)$ of possible values for 
$Y$ (i.e., the set of values over which $Y$ {\em ranges}).  
For simplicity, we assume here that $\V$ is finite, as is $\R(Y)$ for
every endogenous variable $Y \in \V$.
$\F$ associates with each endogenous variable $X \in \V$ a
function denoted $F_X$
(i.e., $F_X = \F(X)$)
such that $F_X: (\times_{U \in \U} \R(U))
\times (\times_{Y \in \V - \{X\}} \R(Y)) \rightarrow \R(X)$.
This mathematical notation just makes precise the fact that 
$F_X$ determines the value of $X$,
given the values of all the other variables in $\U \union \V$.
The structural equations define what happens in the presence of external
interventions. 
Setting the value of some set  $\vec{X}$ of variables to $\vec{x}$
in a causal model $M = (\Scal,\F)$ results in a new causal model, denoted
$M_{\vec{X}\gets \vec{x}}$,
which is identical to $M$, except that the
equations for $\vec{X}$ in $\F$ are replaced by $\vec{X} = \vec{x}$.

Note that the causal models we consider here are deterministic. In
general, one can also consider 
 \emph{probabilistic causal models}.  A probabilistic causal model is a
tuple $M=(\Scal,\F,\Pr)$, where $(\Scal,\F)$ is a causal model, and $\Pr$ is
a probability on contexts. Deterministic models suffice for offering a
qualitative notion of harm, but we use probabilistic causal
models for our quantitative generalization \cite{BCH22b}.

The dependencies between variables in a causal model $M = ((\U,\V,\R),\F)$
can be described using a {\em causal network}\index{causal
  network} (or \emph{causal graph}),
whose nodes are labeled by the endogenous and exogenous variables in
$M$, with one node for each variable in $\U \cup
\V$.  The roots of the graph are (labeled by)
the exogenous variables.  There is a directed edge from  variable $X$
to $Y$ if $Y$ \emph{depends on} $X$; this is the case
if there is some setting of all the variables in 
$\U \union \V$ other than $X$ and $Y$ such that varying the value of
$X$ in that setting results in a variation in the value of $Y$; that
is, there is 
a setting $\vec{z}$ of the variables other than $X$ and $Y$ and values
$x$ and $x'$ of $X$ such that
$F_Y(x,\vec{z}) \ne F_Y(x',\vec{z})$.
A causal model  $M$ is \emph{recursive} (or \emph{acyclic})
if its causal graph is acyclic.
It should be clear that if $M$ is an acyclic  causal model,
then given a \emph{context}, that is, a setting $\vec{u}$ for the
exogenous variables in $\U$, the values of all the other variables are
determined (i.e., there is a unique solution to all the equations).
We can determine these values by starting at the top of the graph and
working our way down.
In this paper, following the literature, we restrict to recursive models.

We call a pair $(M,\vec{u})$ consisting of a causal model $M$ and a
context $\vec{u}$ a \emph{(causal) setting}.
A causal formula $\psi$ is true or false in a setting.
We write $(M,\vec{u}) \sat \psi$  if
the causal formula $\psi$ is true in
the setting $(M,\vec{u})$.
The $\sat$ relation is defined inductively.
$(M,\vec{u}) \sat X = x$ if
the variable $X$ has value $x$
in the unique (since we are dealing with acyclic models) solution
to the equations in
$M$ in context $\vec{u}$ (that is, the
unique vector of values for the exogenous variables that simultaneously satisfies all
equations 
in $M$ 
with the variables in $\U$ set to $\vec{u}$).
Finally, 
$(M,\vec{u}) \sat [\vec{Y} \gets \vec{y}]\varphi$ if 
$(M_{\vec{Y} \gets \vec{y}},\vec{u}) \sat \varphi$.

A standard use of causal models is to define \emph{actual
causation}: that is, 
what it means for some particular event that occurred to cause 
 another particular event. 
There have been a number of definitions of actual causation given
for acyclic models
(e.g., 
\cite{beckers21c,GW07,Hall07,HP01b,Hal48,hitchcock:99,Hitchcock07,Weslake11,Woodward03}).
Although most of what we say in the remainder of the paper applies without
change to other definitions of 
actual causality in causal models, for definiteness, we focus here on
what has been called the \emph{modified} Halpern-Pearl definition
\cite{Hal47,Hal48}, 
which we briefly review.
(See \cite{Hal48} for more intuition and motivation.)

The events that can be causes are arbitrary conjunctions of primitive
events (formulas of the form $X=x$); the events that can be caused are arbitrary
Boolean combinations of primitive events.  
To relate the definition of causality to the (contrastive) definition
of harm, we give a contrastive variant of the definition
of actual causality; rather than defining what it means for
$\vec{X}=\vec{x}$ to be an (actual) cause of $\phi$, we define what it
means for $\vec{X}=\vec{x}$ \emph{rather than $\vec{X} = \vec{x}'$} to be a
cause of $\phi$ rather than $\phi'$. 
\dfn\label{def:AC}
$\vec{X} = \vec{x}$ rather than $\vec{X} = \vec{x}'$ is 
an 
\emph{actual cause} of $\phi$ rather than $\phi'$ in 
$(M,\vec{u})$ if the
following three conditions hold: 
\begin{description}
\vspace{-0.2cm}
\item[{\rm AC1.}]\label{ac1} $(M,\vec{u}) \sat (\vec{X} =
  \vec{x}) \land \phi$.
 \vspace{-0.2cm} 
\item[{\rm AC2.}] There is a set $\vec{W}$ of variables in $\V$
and a setting $\vec{w}$ of the variables in $\vec{W}$
  such that 
$(M,\vec{u}) \sat \vec{W} = \vec{w}$ and
$(M,\vec{u}) \sat [\vec{X} \gets \vec{x}', \vec{W} \gets \vec{w}]\phi'$,
where $\phi' \Rightarrow \neg \phi$ is valid.
\vspace{-0.2cm}
\item[{\rm AC3.}] \label{ac3}\index{AC3}  
  $\vec{X}$ is minimal; there is no strict subset $\vec{X}''$ of
    $\vec{X}$ such that $\vec{X}'' = \vec{x}''$ 
can replace $\vec{X}=\vec{x}'$,
where $\vec{x}''$ is the restriction of
$\vec{x}$ to the variables in $\vec{X}''$.
\end{description}
\edfn
\vspace{-0.2cm}
\noindent AC1 just says that $\vec{X}=\vec{x}$ cannot
be considered a cause of $\phi$ unless both $\vec{X} = \vec{x}$ and
$\phi$ actually happen.  AC3 is a minimality condition, which says
that a cause has no irrelevant conjuncts.  
AC2 captures the standard
but-for condition ($\vec{X}=\vec{x}$ rather than $\vec{X} = \vec{x}'$
is a cause of $\phi$ if, had 
$\vec{X}$ beem $\vec{x}'$ rather than $\vec{x}$, $\phi$
would not have happened) but allows us to apply it while keeping fixed
some variables to the value that they had in the actual 
setting $(M,\vec{u})$.
If $\vec{X} = \vec{x}$ is an actual cause of $\phi$ and there are 
two or more
conjuncts in $\vec{X} = \vec{x}$, one of which is $X=x$,
then $X = x$ is \emph{part of a cause} of $\phi$. 
In the special case that $\vec{W} = \emptyset$, we get the standard
but-for definition of causality:  if $\vec{X} = \vec{x}$ had not
occurred (because $\vec{X}$ was $\vec{x}'$ instead)
$\phi$ would not have occurred (because it would have been $\phi'$).

The reader can easily verify that $\vec{X} = \vec{x}$ is an actual
cause of $\phi$ according to the standard non-contrastive definition
\cite{Hal48} iff there exist $\vec{x}'$ and $\phi'$ such that
$\vec{X} = \vec{x}$ rather than $\vec{X} = \vec{x}'$ is 
an actual cause of $\phi$ rather than $\phi'$ according to our
contrastive definition.

\vspace{-0.3cm}
\section{Defining Harm}\label{sec:harmdef}
\vspace{-0.1cm}
Many definitions of harm have been considered in the literature.  The
ones most relevant to us are those involving causality and
counterfactuals, which have been split into two groups, called the
\emph{causal account of harm} and the \emph{counterfactual comparative
account account of harm}.  Carlson et al.~\citeyear{CJR21} discuss many
variants of the causal account; they all have the following structure:

\textit{
An event $e$ harms an agent {\bf ag} if and only if there is a state
of affairs $s$ such that (i) $e$ causes $s$ to obtain, and (ii) $s$ is a harm
for {\bf ag}.
}

The definitions differ in how they interpret the second clause.
We note that although these definitions use the word ``cause'', it is
never defined formally.  ``Harm'' is also not always defined, although
in some cases the second clause is replaced by phrases that
are intended to be easier to interpret.
For example,
what Suits \citeyear{Suits01} calls the \emph{causal-intrinsic
badness account}  takes $s$ to be a harm for {\bf ag} if $s$ is
``intrinsically bad'' for {\bf ag}.

The causal-counterfactual account (see, e.g.,
\cite{Gar15,Nor15,Tho11, Gar15})
has the same
structure; the first clause is the same, but now 
the second clause is replaced by a phrase involving
counterfactuals.  In its simplest version, this can be formulated as
follows: $s$ is a harm for {\bf ag} if and only if {\bf ag} would have
been better off had $s$ not obtained.

Even closer to our account is what has been called
the \emph{contrastive causal-counterfactual account}.  For example,
Bontly \citeyear{Bontly16} proposed the following:

\textit{
An event $e$ harms a person {\bf ag} if and only if there is a state
of affairs $s$ and a contrast state of affairs $s'$ such that (i) $e$
rather than a 
contrast event $e'$ causes $s$ rather than $s'$ to obtain, and (ii)
{\bf ag} is worse off in $s$ than in $s'$.
}

Our formal definition of harm is quite close to Bontly's.  We replace
``state of affairs'' by ``outcomes'', and associate with each outcome
a utility.  This is essentially the standard model in decision theory, 
where actions map states to outcomes, which have associated utilities.
Besides allowing us to connect our view to the standard
decision-theoretic view (see, e.g., \cite{res,Savage}),
this choice means that we can benefit from
all the work done on utility by decision theorists.

To define harm formally in our framework, we need to both extend and
specialize causal models: 
    We specialize causal models by assuming that they include a
    special 
    endogenous 
    variable $O$ for \emph{outcome}.  The various values of
    the outcome value will be assigned a utility. We
    often think of an action as affecting many variables, whose values
    together constitute the outcome.  The decision to ``package up''
        all these variables into a single variable $O$ here is deliberate;
    we do not want to consider the causal impact of some variables
    that make up the outcome on other variables that make up the
    outcome (and so do not want to allow interventions on individual
    variables that make up an outcome; we allow only interventions on
    complete outcomes).
    On the other hand, we extend causal models by assigning  a utility
value to outcomes (i.e., on 
values of the outcome variable), and by having a default utility.

We thus take a \emph{causal utility model} to be one of the form 
$M= ((\U,\V,\R),\F, {\bf u}, d)$, where $(\U,\V,\R),\F)$ is a causal model one
of whose endogenous variables is $O$, 
${\bf u}:\R(O) \rightarrow [0,1]$ is a
utility function on outcomes (for
simplicity, we assume that utilities are normalized so that the best
utility is 1 and the worst utility is 0), and $d \in [0,1]$ is a
default utility.%
\footnote{As we said in the introduction, in general, we think of the default
utility as being context-dependent, so we really want a function from
contexts to default utilities.  However, in all the examples we
consider in this paper, a single default utility suffices, so for ease
of exposition, we make this simplification here.}
As before, we call a pair $(M,\vec{u})$, where now
$M$ is a causal utility model and  $\vec{u} \in \R(\U)$, a setting.  

Just like causality, we define harm relative to a setting.  
Whether or not an event $\vec{X} = \vec{x}$ harms an agent in a given
setting will depend very much on the choice of utility function and
default value.  Thus, to justify a particular ascription of harm, we
will have to justify 
both 
these choices.  
In the examples we consider, we typically view the utility
function to be {\bf ag}'s utility function, but we are not
committed to this 
choice (\eg when deciding whether harm is caused by a parent not
giving a child ice cream, we may use the parent's definition of utility,
rather than the child's one).
%
The choice of a default value is more complicated, and will be discussed 
when we get to examples; for the definition itself, we assume that we
are just given the model,
including utility function and default value.

The second clause
of our definition is a formalization of Bontly's
definition, using the definition of causality given in Section~\ref{sec:models},
where the events for us, as in standard causal models,
have the form $\vec{X}=\vec{x}$ and the alternative events have the form
$\vec{X} = \vec{x}'$, and they cause outcomes $O=o$ and $O=o'$, respectively.
Unlike Bontly's definition (and others), not
only do we require that {\bf ag} is worse off in outcome 
$o$ (the analogue of state of affairs $s$) than
in outcome $o'$ (where ``worse off'' is formalized by taking the utility to be
lower), we also require the utility of $o$ to be lower than the
{\em default utility}.
There is also an issue as to whether we consider there to be harm if
$\vec{X} = \vec{x}'$ results in a worse outcome than $o$.  Since
intuitions may differ here, we formalize this requirement in a third
clause, H3, and use it to distinguish between harm and \emph{strict} harm.
We will see the effects of our modifications to Bontly's definition
when we consider examples in Section~\ref{sec:exam}.  
  \vspace{-0.1cm}
\dfn\label{def:harm2}
$\vec{X} = \vec{x}$ 
\emph{harms} {\bf ag} in $(M,\vec{u})$, where $M = ((\U,\V,\R),\F, {\bf u},d)$, 
if there exist $o\in \R(O)$ and $\vec{x}' \in \R(\vec{X})$ such that
\begin{description}
  \vspace{-0.2cm}
  \item[{\rm H1.}]\label{h1} ${\bf u}(O=o) < d$; and
    \vspace{-0.2cm}
\item[{\rm H2.}]\label{h2} there exists $o' \in \R(O)$ such that
  $\vec{X}=\vec{x}$ rather than $\vec{X}=\vec{x}'$ causes $O=o$ rather
  than $O=o'$ and  
  ${\bf u}(O=o) < {\bf u}(O=o')$.
  \end{description}
$\vec{X} = \vec{x}$  \emph{strictly harms} {\bf ag} in $(M,\vec{u})$
if, in addition,
\begin{description}
\item[{\rm H3.}]\label{h3}  
   ${\bf u}(O=o) \le {\bf u}(O=o'')$ for the unique $o'' \in \R(O)$
such that $(M,\vec{u}) \sat [\vec{X} \gets \vec{x}']( O=o'')$. 
\end{description}
\edfn
  \vspace{-0.2cm}
In the special case where Definition \ref{def:harm2} is satisfied for some value $o'$ appearing in H2 such that ${\bf u}(O=o) < d \leq {\bf u}(O=o')$, we say that {\em $\vec{X}=\vec{x}$ causes {\bf ag}'s utility to be lower than the default}. 

It is important to point out that it is quite rare for harm and strict
harm to come apart. For one, it requires causation and but-for
causation to come apart (otherwise $o' = o''$).  In
addition, it requires $O$ to have at least three values (otherwise
again $o'=o''$). Lastly, even if these two conditions are met, we
also need that ${\bf u}(O=o'') < {\bf u}(O=o) < {\bf u}(O=o')$. We
discuss one example in the supplementary material (see Section
\ref{sec:h3}) in which these conditions are met.

 As with most concepts in actual causality, deciding whether harm
 occurred is intractable. Indeed, it 
 is easy to see that it is at least as hard as causality, which is
  DP-complete \cite{Hal47}. 
 However,
 this is unlikely to be a problem in
 practice, since we expect that  the causal models that arise when we
 want to deal with harm will have few variables, which take on few
 possible values (or will involve many individuals that can all be
 described with by a small causal model), so we can
  decide harm by simply checking all possibilities.
 
  It is useful to 
compare 
our definition with the
counterfactual comparative
account of harm.  Here it is,
translated into our 
notation: 
\vspace{-0.1cm}
\dfn\label{def:harm3}
$\vec{X} = \vec{x}$ 
\emph{counterfactually harms} {\bf ag}
in $(M,\vec{u})$, where $M = ((\U,\V,\R),\F, {\bf u},d)$ 
 if there exist $o, o' \in \R(O)$ and $\vec{x}' \in \R(\vec{X})$ such that
\begin{description}
\vspace{-0.2cm}
\item[{\rm C1.}]\label{c1} $(M,\vec{u}) \sat \vec{X}
  = \vec{x} \land O=o$;  
  \vspace{-0.2cm}
\item[{\rm C2.}] $(M,\vec{u}) \sat [\vec{X} \gets \vec{x}']( O=o')$;
\vspace{-0.2cm}
\item[{\rm C3.}] \label{c3}\index{C3} ${\bf u}(O=o) < {\bf u}(O=o')$.
\end{description}
\edfn
\vspace{-0.2cm}
That is, $\vec{X} = \vec{x}$ counterfactually harms {\bf ag} if, for
some $x'$ and $o'$, 
$\vec{X}=\vec{x}$ is what actually happens (C1), $O=o'$ would have
happened had $\vec{X}$ been set to $\vec{x}'$ (C2), and {\bf ag} gets
higher utility from $o'$ than from $o$ (C3).
C1 and C2 together are equivalent to AC1 and AC2 in the special case that 
$\vec{W} = \emptyset$.  That is, C1 and C2 essentially amount to
but-for causality.  C3 differs from 
our conditions
 by not taking into account the
default value.  

Note that  
Definition \ref{def:harm3}
has no analogue of AC3, but all the examples
focus on cases where $\vec{X}$ is actually a singleton, so AC3 is
trivially satisfied. 
 The key point from our perspective is that the 
 counterfactual comparative account
considers 
only but-for causality, and does not
consider a default value.  The examples in the next section show how
critical these distinctions are.

As mentioned 
earlier, RBT recently developed
a formal account of harm using causal models.
While their account is probabilistic and quantitative, we can consider
the special case 
where everything is deterministic and qualitative. 
When we do this, their account
reduces to a
strengthening of Definition \ref{def:harm3} that brings it somewhat
closer to our account: they also suggest using defaults, but have
default actions rather than default utilities.   In their version of
Definition~\ref{def:harm3}, $\vec{X}$ is taken to be the variable
representing the action(s) performed and $x'$ is the default action.
\commentout{
Finally, as mentioned earlier, they also use causality, but their
account differs from the standard account \cite{Hal48} in some significant
ways; see Section~\ref{sec:pre}.
}
In order to deal with the limitations of but-for causality, RBT offer
a more general account (see their Appendix A) that 
uses path-specific causality, instead of actual causation. This makes their account different from ours in some significant 
respects; see Section~\ref{sec:RBT}.
  
\vspace{-0.3cm}
\section{Examples}\label{sec:exam}
\vspace{-0.1cm}
We now analyse several examples to illustrate how our definition
handles the most prominent issues that have been raised in the
literature on harm.
Bradley \citeyear[p. 398]{Bra12} identifies two such issues that strike him
``as very serious'', 
namely the problem of preemption, and the problem of
distinguishing harm from merely failing to benefit. These problems
therefore serve as a good starting point.  

\vspace{-0.2cm}
\subsection{Preemption}\label{sec:pre}
\vspace{-0.1cm}
To anyone familiar with the literature on actual causation what
follows will not come as a surprise. Lewis 
used examples of
preemption to argue 
 that there can be causation without counterfactual dependence
(i.e., we need to go beyond but-for causality);
this conclusion is now universally accepted.
Essentially the same examples show up in the literature on harm:
cases of preemption show that an event can cause harm even though the
agent's well-being does not counterfactually depend on it.
Thus, the counterfactual comparative account of harm fails for the
same reason it failed for causality.  The good news is that 
the formal definition of causation (by design) handles problems like
preemption well; moreover,
the solution carries over directly to our definition of harm.  
The following vignette is due to Bradley \citeyear{Bra12}, but issues of preemption show up in many papers on 
causality~\cite{beckers21c,Hall07,HP01b,Hal48,Hitchcock07,Weslake11};
all can be dealt with essentially the same way.

\begin{example}[Late Preemption]\label{ex:lp}
\emph{Suppose Batman drops dead of a heart attack. A millisecond after his
death, his body is hit by a flaming cannonball. The cannonball would
have killed Batman if he had still been alive. So the counterfactual
account entails that the 
heart attack was not harmful to Batman. It didn't make things go worse
for him. But intuitively, the heart attack was harmful. The fact that
he would have been harmed by the flaming cannonball anyway 
does not seem relevant to whether the heart attack was actually
harmful.
}

\emph{In terms of the formal definition, we take $H$ to represent
whether Batman has a heart attack ($H=0$ if he doesn't; $H=1$ if he
does), $C$ to represent if Batman is hit by a cannonball, and 
$D$ to represent whether Batman dies.  Let $\vec{u}$ be the context
where $H=1$.    Even without describing the equations, according to
the story, $(M,\vec{u}) \sat H=1 \land D=1 \land [H=0](D=1)$: Batman
has a heart attack and he dies, but he would have died even if he did
not have a heart attack (since he would have been hit by the cannon
ball).  Thus, C3 does not hold, since $o=o'$; the outcome is the same
whether or not Batman has a heart attack.
}

\emph{
The standard causal account handles this problem by introducing 
two new variables: $K$, for ``Batman is killed by the cannonball'',
and $S$, for ``Batman died of a heart attack'',
to take into account the temporal asymmetry between death due to a heart attack
and death due to a cannonball.
(We could also deal with this asymmetry by having ``time-stamped''
variables that talk about when Batman is alive. For more details on
incorporating temporal information by using time-stamped variables,
see \cite{Hal48}.)   
The causal model has the following equations:
\commentout{
\begin{itemize}
\item $D =  S \lor K$ (i.e., $D=1$ if either $S=1$ or $K=1$: Batman
dies if his heart stops or the canonball kills him);
\item  $S=H$ (Batman's heart stops if he has a heart attack); 
\item $K=\lnot S \land C$ (Batman is killed by the canonball if the
  canonball hits him and 
  his heart is still beating).
\end{itemize}
}
}
%
\emph{$D =  S \lor K$ (i.e., $D=1$ if either $S=1$ or $K=1$: Batman
dies if he has a heart attack or the canonball kills him);
$S=H$ (Batman's heart stops if he has a heart attack); and
 $K=\lnot S \land C$ (Batman is killed by the canonball if the
  canonball hits him and 
  his heart is still beating).
  }
%
\emph{We now get that Batman's heart attack 
rather than 
its absence
 is a cause of
him being alive rather than dead.  Clearly $(M,\vec{u}) \sat H=1 \land
D=1$.  If we fix $K=0$ (its actual value, since the cannonball in
fact does not kill Batman; he is already dead by the time the
cannonball hits him), then we have that $(M,\vec{u}) \sat [H=0,
K=0](D=0)$, so AC2 holds.   Thus, 
the causal part of H2
holds.
(See \cite[Example 2.3.3]{Hal48} for a detailed discussion of an isomorphic example.)}

\emph{If we further assume, quite reasonably, that Batman prefers being
alive to being dead 
(so the utility of being alive is higher than that of being dead)
and that the default utility is that of him being alive, then 
H1 and H2 hold.
Thus, our definition of harm avoids
the counterintuitive conclusion by observing that Batman's heart
attack caused his death, thereby causing the utility to be lower than
the default.} \hfill \wbox
\end{example}
\vspace{-0.2cm}
Our analysis of preemption is indicative of the more general point
that many of the issues plaguing the literature on harm can be
resolved by making use of causal models and the definitions of
causation that they allow. Causal models allow a more precise and
explicit representation of the relevant causal structure, thereby
forcing a modeler to make modeling choices that resolve the inherent
ambiguity that comes with an informal and underspecified causal
scenario. Obviously such modeling choices can be the subject of
debate (see \cite{HH10} for a discussion of these modeling choices).
The point is not that using causal models by itself determines
a unique verdict on whether harm has occurred, but rather that such a
debate {\em cannot even be had} without being explicit about the
underlying causal structure.

\vspace{-0.2cm}
\subsection{Failing to Benefit}\label{sec:fail}
\vspace{-0.1cm}
%
One of the central challenges in defining harm is to distinguish it
from merely failing to benefit. 
Although most authors define benefit simply as the symmetric
counterpart to harm, we do not believe that this is always
appropriate; we return to this issue in \cite{BCH22b} where we
consider more quantitative notions of harm.  But for the current
discussion, we can set this issue aside:
what matters is that merely failing to make someone better off does
not in itself suffice to say that there was harm. Carlson et
al.~\citeyear{CJR21} present the 
following well-known scenario to illustrate the point.



\begin{example}[{\bf Golf Clubs}]\label{xam:golf}
\emph{Batman contemplates giving a set of golf clubs to Robin, but
eventually decides to keep them. If he had not decided to keep them,
he would have given the clubs to Robin, which would have made Robin
better off.} 

\emph{By keeping the golf clubs, Batman 
clearly failed to make Robin
better off.  
The counterfactual account
considers any such failure 
to result in harm.
Indeed, it is easy to see that C1--C3 hold.  If we take $GGC$ to
represent whether Batman gives the golf clubs to Robin ($GGC=1$ if he
does; $GGC=0$ if he doesn't) and the outcome $O$ to represent whether
Robin gets the golf clubs ($O=1$ if he does; $O=0$ if he doesn't),
then $GGC=0$ is a but-for cause of $GGC=0$, so C1 and C2 hold.  If we
further assume that Robin's utility of getting the golf clubs is
higher than his utility of not getting them, then C3 holds.
Yet it 
sounds
 counterintuitive to claim that Batman harmed Robin on
this occasion.} \hfill \wbox 
\end{example}
\vspace{-0.2cm}
Although 
H2
 holds in our account of harm (for the same reason that C1--C3
hold), we avoid the counterintuitive conclusion by assuming that the
default utility is ${\bf u}(O=0)$, so 
H1
 does not hold.  This seems to us reasonable; there is nothing
in the story that suggests that Robin is entitled to expect golf
clubs.  On the other hand, if we learn that 
Batman is a professional golfer, Robin has been his reliable caddy
for many years, and that at the start of every past season Batman has 
purchased a set of golf clubs for Robin, then it sounds quite
plausible that the default is for Robin to 
receive
 a set of golf clubs.
With this default, 
H1
 does hold, and 
our definition concludes that Robin {\em has} been
harmed. Thus our account can offer different verdicts depending on the
choice of 
default utility. As we said in the introduction, we 
view this flexibility as a feature of our account.
This point is highlighted in the following, arguably more realistic,
scenario. 
(RBT make exactly the same point as we do when they analyze
such examples 
\cite[p. 15]{RBT22}.)  

\begin{example}[{\bf Tip}]\label{xam:tip}
\emph{Batman contemplates giving a tip to his waiter, but eventually decides to keep the extra money for himself. If he had not decided to keep it, he would have given it to the waiter, which would have made the waiter better off.}

\emph{To those living in the US, it does not at all sound counterintuitive
to claim that Batman harmed the waiter, for his income substantially
depends on receiving tips and he almost always does receive a tip.  
Indeed, if we take the default utility to be that of receiving a tip,
then in this example, the waiter is 
 harmed by Batman not giving
a tip.  By way of contrast, in countries in Europe where a tip would
not be expected, it seems to us reasonable to take the default utility to be
that of not receiving a tip.  In this case, the waiter would not be
harmed.}  
\hfill \wbox
\end{example}
\vspace{-0.2cm}


Examples \ref{xam:golf} and \ref{xam:tip} are isomorphic as far as the
causal structure goes; we can take the utilities to be the same as
well.  This means that we need additional structure to be able to
claim that 
the agent is harmed
 in one case and not the other.  That additional structure in
our framework, which we would argue is quite natural, is the choice of
default utility. 
Note that neither scenario
explicitly mentions what the default utility should be. We thus need to
rely on further background information to make a case for a particular
choice.  
There can be many factors that go into
determining
 a good default. We therefore  do not give a general recipe for
doing so.  Indeed, as we pointed out in the introduction with the
euthanasia example, reasonable people can disagree about the
appropriate default (and thus reach different conclusions regarding harm).

\vspace{-0.2cm}
\subsection{Preventing Worse}
\vspace{-0.1cm}

There exist situations in which the actual event rather than an
alternative event causes a bad outcome rather than a good outcome, but
the alternative results in an even worse outcome.
Because of the latter, we do not consider these situations to be cases of
strict harm, due to condition H3 in Definition \ref{def:harm2}.
From the perspective of the car
manufacturer, this is precisely what is going on in our starting
Example \ref{ex:car}, but Bob might disagree. We now take a closer
look at this example to bring out the conflicting perspectives. 
\vspace{-0.1cm}
\begin{example}[Autonomous Car]\label{ex:AC} \emph{Let $O$ be a three-valued variable capturing the outcome for Bob, with
the utility defined as equal to the value of $O$. 
$O=0.5$ stands for the injury resulting from crashing into the safety fence, 
and a potentially more severe injury resulting from crashing into the stationary car is captured by $O=0$. Bob not being injured is $O=1$.}

\emph{Recall that the system 
has the built-in standard that the driver's reaction time is $10$ seconds, which is too long to avoid colliding into the stationary car. Imagine the manufacturer implemented this standard by restricting the system's actions in such cases to two possibilities: do not intervene ($F=0$) or drive into the fence ($F=1$). This means that the causal structure is very similar to our Late Preemption example (Example \ref{ex:lp}), for hitting the fence preempts the collision with the stationary car. We therefore add a variable to capture the asymmetry between hitting the fence and hitting the stationary car: $FH$ and $CH$ respectively. The equation for $O$ is then such that $O=1$ if $FH=CH=0$, $O=0.5$ if $FH=1$, and $O=0$ if $CH=1$ and $FH=0$.}

\begin{figure}[h]
  \begin{center}
    \begin{tikzpicture}[line cap=round,line join=round,x=1cm,y=1cm,scale=2.3]
    \clip(1,0.4) rectangle (2.5,1.65);
\draw [-stealth,line width=1pt] (2.2,1.5) -- (1.5,1.5);
\draw [-stealth,line width=1pt] (1.5,1.5) -- (1.5,1);
\draw [-stealth,line width=1pt] (2.2,1.5) -- (2.2,1);
\draw [-stealth,line width=1pt] (1.5,1) -- (2.2,1);
\draw [->,line width=1pt] (1.5,1) -- (1.8,0.5);
\draw [->,line width=1pt] (2.2,1) -- (1.86,0.5);
\begin{scriptsize}
\draw [fill=black] (1.5,1.5) circle (0.9pt);
\draw[color=black] (1.55,1.6) node {$F$};
\draw [fill=black] (2.2,1.5) circle (0.9pt);
\draw[color=black] (2.24,1.6) node {$C$};
\draw [fill=black] (1.5,1) circle (0.9pt);
\draw[color=black] (1.63,1.13) node {$FH$};
\draw [fill=black] (2.2,1) circle (0.9pt);
\draw[color=black] (2.35,1.13) node {$CH$};
\draw [fill=black] (1.83,0.49) circle (0.9pt);
\draw[color=black] (1.96,0.5) node {$O$};
\end{scriptsize}
\end{tikzpicture}
\caption{Causal graph for Ex.~\ref{ex:AC}}
\end{center}
  \end{figure}
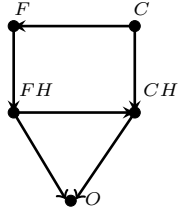

\emph{As the autonomous car drives towards the fence only because there is a stationary car, the equation for $F$ is $F=C$ (where $C$ represents the presence of the car). The fact that hitting the fence prevents hitting the car is captured in the equation for $CH$: $CH=C \land \lnot FH$. Lastly, we have $FH=F$. The context is such that $F=1$ and $C=1$, and thus $FH=1$, $O=0.5$, and $CH=0$. 
}

\emph{Did the system harm Bob? Carlson et al.~\citeyear{CJR21} answer
this in the 
negative for an example that is essentially the same as this one (see
their ``Many Threats'' example), and use this verdict to argue
against the causal-counterfactual account, which reaches the opposite
verdict. According to them, the contrastive causal-counterfactual
account does not reach this verdict because there is
no contrastive causation here, but as they do not give a definition
of causation, it is impossible to reconstruct how they arrive at this
verdict.  In any case, what matters for our purposes is that our
definition of causation does consider the system's hitting the fence
rather than not intervening to cause Bob being mildly injured rather
than not being injured at all. To see why, observe that taking
$\vec{W}$ to be $CH$, we get that $F=1$ rather than $F=0$ causes
$O=0.5$ rather than $O=1$: $(M,\vec{u}) \sat [F \gets 0, CH \gets
  0]O=1$.  
}

\emph{Therefore, if we assume 
that the
default utility is that of Bob not being injured, conditions H1 and H2
are satisfied and there is harm. Notice though that $F=1$ rather than
$F=0$ is also a 
but-for cause of $O=0.5$ rather than $O=0$, that is, Bob's being mildly
injured rather than severely injured counterfactually depends on the
system's action. This is where condition H3 kicks in: it ensures that
we do not consider there to be strict harm caused if the alternative
would have resulted in an even worse outcome.
Thus, the car manufacturer could make the case that although their
policy harmed Bob, it was justified in doing so. 
}

\emph{More generally, it is an easy consequence of our definitions that in cases where $\vec{X}=\vec{x}$ rather than $\vec{X}=\vec{x}'$ causes harm but not strict harm, the alternative event {\em would} have resulted in strict harm, i.e., $\vec{X}=\vec{x}'$ rather than $\vec{X}=\vec{x}$ would have caused strict harm. As a result, it is sensible in such cases for someone to argue that they were justified in causing harm, as the alternative would have been worse.
}

\emph{Bob, on the other hand, believes he has been 
strictly
harmed, because he claims that he could have prevented the collision if he had been alerted. This disagreement can be captured formally by stating that Bob is using a three-valued variable $F$ instead of a binary one, where the third option ($F=2$) corresponds to alerting Bob. Incorporating this variable into the model (and assuming that Bob is correct regarding his driving skills) we would again get that $F=1$ rather than $F=2$ causes $O=0.5$ rather than $O=1$, but with the important distinction that H3 {\em is} satisfied for these contrast values and thus the system's action does 
strictly 
harm Bob. Our analysis does not resolve the conflict (and it is not meant to do so), instead it allows for a precise formulation of the source of the disagreement.
}
\hfill \wbox
\end{example}
\vspace{-0.2cm}

%
In the supplementary material, we describe four more examples
illustrating multiple contrasts for the outcome, the role of the choice and the range of variables and the choice of default, and our rationale for considering a contrastive definition, rather than causal-counterfactual one.

\section{Comparison to RBT}\label{sec:RBT}
In this section, we do a more careful comparison of our approach and
that of RBT.  RBT focus on choices made by an agent, where these
are choices of what action to take, and assume that there is a default
action, to which they compare the choice made by the agent.
It follows easily from RBT's definition that if the agent 
performs the default action, there is no harm. Yet
there are many instances in which doing what is morally preferable
causes harm, albeit accidentally. Simply imagine a doctor prescribing
medication to a patient, and the patient unfortunately suffering a
very rare allergic reaction to the medication, where the reaction is
far worse than the initial condition that the patient had. Then
clearly the doctor harmed the patient. The most obvious choice of
default action here is the actual action
(and, in fact, RBT themselves mention following ``clinical guidelines''
as an example of a default action in their Appendix C). But this means that
according to RBT's definition there would not be harm here.  Although
we use a default utility, there is no need for this default utility to
be the utility of a default action, so we do not have this problem.

Another significant difference between our approach and that of RBT is
that, although RBT use causal models, unlike us, they do not use 
a sophisticated definition of actual causality such as the one
given by Halpern \citeyear{Hal47,Hal48}.  
In their Definition 3, RBT consider but-for causality.  Not surprisingly,
this will not suffice to deal with problematic examples where the more
general notion of causality is needed (see, e.g.,
Example~\ref{ex:lp}).  In their Definition 9, they generalize
Definition 3 to allow for path-dependent causality.  They do not
explain how they choose which paths to consider, but 
in both Example~\ref{ex:lp} and the corresponding example
of late preemption they consider, by choosing the appropriate paths,
they can simulate the effects of AC2.   (Specifically, they can
simulate the effect of choosing a 
set $\vec{W}$ of variables and fixing them to their actual values.)  As
a consequence, they 
get the same results as those obtained by 
Halpern's definition of
actual causality.  It is not clear whether this will always be the
case.   More importantly, the ability to determine harm relative to
some choice of paths gives the modeler a significant extra degree of
freedom to tailor the results obtained.
We believe that if paths are going to be used, there needs to be a
more principled analysis of how to go about choosing them.

\vspace{-0.3cm}
\section{Conclusion}\label{sec:con}
\vspace{-0.1cm}
We have defined a qualitative notion of harm, and shown that it deals
well with the many problematic examples in the philosophy literature.
We believe that our definition will be 
widely
 applicable in 
the 
 regulatory frameworks that 
we expect
 to be designed soon
in order
 to deal with autonomous systems.

Of course, having a qualitative notion of harm is only a first step.
For practical applications, we often need to quantify harm; for
example, we may want to choose the least harmful of a set of possible
interventions. As we said, we develop a quantitative notion of harm in
\cite{BCH22b}.  
While one could just define a quantitive notion that considers the difference between the
utility of the actual outcome and the default utility (this is
essentially what RBT do), we believe that the 
actual problem is more
nuanced.  For example, even if we can agree on the degree of harm to
an individual, if there are many people involved and there is
a probability of each one being harmed, should we just sum the
individual harms, weighted by the probability?  We argue that this is not always appropriate, and 
discuss alternatives, drawing on work from the decision-theory literature.

\section*{Acknowledgements}

The authors would like to thank the NeurIPS reviewers for their
detailed comments and Jonathan Richens for a fruitful discussion of a
preliminary version of this paper. Sander Beckers was supported by the
German Research Foundation (DFG)
under Germany’s Excellence Strategy – EXC number 2064/1 – Project number
 390727645, and by the Alexander von Humboldt Foundation.
Hana Chockler was supported in part by the UKRI Trust-worthy Autonomous Systems Hub (EP/V00784X/1)
and the UKRI Strategic Priorities Fund to the UKRI Research Node on Trustworthy Autonomous Systems Governance and Regulation (EP/V026607/1).
Joe Halpern was supported in part by NSF grant IIS-1703846, 
ARO grant W911NF-22-1-0061, and MURI grant W911NF-19-1-0217.

\bibliographystyle{plainnat}
\bibliography{fv,joe,z,allpapers}

\appendix

\section{Appendix}

Here we provide some additional details on three topics. First,
we illustrate the role played by H3 in our definition of harm. Second,
we discuss in more detail
how our approach differs from that of RBT. Third, we
present four more examples that illustrate how our definition handles
issues which have been discussed in the harm literature. 

\subsection{Discussion of H3}\label{sec:h3}

As we  mentioned, H3 is intended to capture the intuition that
there is no strict harm if the alternative would have resulted in an
even worse outcome. For example, following the 
reasoning of the 
car manufacturer in Example \ref{ex:AC}, the system's decision to
drive into the fence rather than doing nothing is not 
strictly 
harmful because
Bob would have suffered even worse injuries had the system done nothing.
Since H1 and H2 are satisfied for this particular
contrastive event, our definition would reach the opposite verdict if
it weren't for H3. Note that the counterfactual comparative account (Definition
\ref{def:harm3}) also says that there is no harm:
the alternative event under consideration would have
given a worse outcome, so that C3 is not satisfied, and therefore
there is no harm.
Considering H3 gives more insight into the differences between
the counterfactual account and ours.

Suppose that we consider some contrastive event $\vec{X}=\vec{x}'$
such that $(M,\vec{u}) \sat \vec{X} = \vec{x} \land O=o$ and
$(M,\vec{u}) \sat [\vec{X} \gets \vec{x}']( O=o'')$,
so C1 and C2 hold, and the first half of H2 holds if $o'' \ne o$:
$\vec{X} = \vec{x}$ 
rather than $\vec{X} = \vec{x}'$ causes $O=o$ rather than $O=o''$.
H3 plays no role if H1 is not satisfied, so for simplicity, suppose
that H1 also holds. 
Then it is easy to see that whenever ${\bf u}(O=o) \neq {\bf
  u}(O=o'')$, our definition
  of strict harm
gives the same verdict as the counterfactual comparative definition
for this particular contrast (i.e., for this choice of $\vec{x}'$):
if ${\bf u}(O=o) < {\bf u}(O= o'')$, then $o'' \ne o$, 
so H2 holds, 
as do H3 and C3; it follows that
according to
 both
definitions $\vec{X} = \vec{x}$ 
harms
 the agent.  On the other hand, if 
${\bf u}(O=o) > {\bf u}(O= o'')$, then neither C3 nor H3 hold (for
this choice of $\vec{x}'$).


What happens if ${\bf u}(O=o) = {\bf u}(O=o'')$?
 This can happen for two reasons:
\begin{enumerate}
  \item there is no but-for causation, that is, $o=o''$;
\item there is but-for causation but the counterfactual outcome
    $O=o''$ happens to have  utility identical to the actual outcome. 
\end{enumerate}
Thus, roughly speaking (and ignoring the key role played by the
default utility), our definition differs from the counterfactual
comparative account only if
$\vec{X}=\vec{x}$ rather than $\vec{X}=\vec{x}''$ is not a but-for
cause of the actual utility: changing $\vec{x}$ into $\vec{x}''$ does
not change the agent's utility.

Examples in
which the first reason is relevant are widespread and crucial to our
analysis, for those are precisely the examples in which actual
causation (Definition \ref{def:AC}) and but-for causation come
apart. Our Late Preemption example (Example~\ref{ex:lp}) offers one
illustration, the literature on actual causation contains many more.
An example where the second reason is relevant involves a more subtle
way in which but-for causation comes apart from actual causation.
Consider a ``Sophie's choice'' like setting: An agent must choose
whether $X=1$ or $X=2$.  There are two children, who will either live
or die depending on the choice:  if $X=i$ is chosen,
then child $i$ lives ($L_i = 1$) and child 
$3-i$ dies ($L_{3-i} = 0$).
The possible outcomes are that both children live ($o_{11}$), just
child 1 lives ($o_{10}$), just child 2 lives ($o_{01}$), and neither
child lives ($o_{00}$), where $d= {\bf u}(O = o_{11}) >
{\bf u}(O = o_{10}) = {\bf u}(O = o_{01}) > {\bf u}(O = o_{00})$.
In fact, $X=1$ is chosen, so we get but-for causality,
but switching from $X=1$ to $X=2$ gives an outcome of equal utility.
However, if we hold $L_1 = 1$ fixed (which we can do in our framework
to show causality) and switch to $X=2$, then we get the outcome $O =
o_{11}$.  Thus, in our framework $X=1$
strictly 
 harms the agent; in the causal
counterfactual framework, it does not.

This emphasizes the point we (and RBT) made 
that one set of problems
that occur in defining harm is identical to the type of problems that
occur in defining causation, and can be solved in the same way.


\subsection{More Examples}
We present four further examples that illustrate how our approach deals
with the difficulties of defining harm that have been highlighted in
the literature. 

The cases in Sections \ref{sec:pre} and \ref{sec:fail} all involved a
binary outcome; there were 
only two relevant events that could occur.
Carlson et
al.~\citeyear{CJR21} 
discuss cases that involve more than two possible events in order to
argue against existing causal accounts. 
The following example
forms one instance of their argument. 



\begin{example}[{\bf Tear Gas}]
\emph{The Joker sprays tear gas in exactly one of Batman’s eyes. If he
had not done that, he would have sprayed tear gas in both of Batman’s
eyes, which would have made Batman even worse off. One of the
alternatives available to the Joker, however, was to simply leave
Batman alone.} 

\emph{Intuitively here Joker harms Batman when he sprays him. 
To argue that the ``incorrect'' answer is obtained by the definition
of harm they focus on, Carlson et al.~consider a specific
alternative event, namely, that Joker sprays tear gas in both of
Batman's eyes, while observing that other alternatives (like leaving
Batman alone) are also available.
Rather than existentially quantifying over $\vec{x}'$, as we have done,
(both 
in Definition~\ref{def:harm2} and the gloss of the counterfactual harm
definition given in Definition~\ref{def:harm3}), they take a version of
counterfactual harm where $\vec{X} = \vec{x}'$ is taken 
to be the closest alternative to
$\vec{X}=\vec{x}$ (according to some implicit, but unspecified, notion
of closeness).  Both our definition of harm and our gloss of the
counterfactual definition (with the obvious assumptions about utility,
and taking the default utility to be that of Batman being unharmed for
our definition) agree that Joker did harm Batman in this case, as we
would expect.}

\emph{In this example, there are three events of interest (Joker sprays tear
gas in one eye; Joker sprays tear gas in both eyes; Joker doesn't
spray tear gas at all).  We can model this using a variable $TG$ that
takes on three possible values (say, $0$, $1$, and $2$).  According to
Definition~\ref{def:harm3}, as long as one of them leads to a better
utility than what actually happened, there was harm.  But as the golf
clubs example shows, this conclusion is not always justified; in
general, we need to take defaults into account.} \hfill \wbox
\end{example}
\vspace{-0.2cm}
\commentout{
We present two further examples that illustrate how our approach deals
with the difficulties of defining harm that have been highlighted in
the literature.
The first example, due to Shiffrin \citeyear{Shiffrin99}
illustrates the role of both the choice of the
range of variables in the causal model and the choice of default.
}

Now we present an example, due to Shiffrin \citeyear{Shiffrin99},
that illustrates the role of both the choice of the
range of variables in the causal model and the choice of default.

\begin{example} \emph{Betty is drowning in a fast-moving river.  Veronica
rescues her by grabbing her arm and pulling her out, accidentally
fracturing Betty's humerus.}

\emph{Did Veronica's rescue 
harm Betty?
Shiffrin claims it
does because Veronica could have pulled her out without breaking her
arm.  Indeed, Klocksiem \citeyear{Klock12}, in his analysis, points
out that ``it seems possible to rescue someone from drowning
without breaking her arm''.  The first step in our analysis is to
decide whether we should allow this possibility.  That is, suppose
that we have a variable $P$ that describes how and whether Veronica
pulls out Betty.  We can take $P=0$ if Veronica does not pull out
Betty, $P=1$ if she pulls her out by 
grabbing (and breaking) her arm.
The modeler must then decide whether to allow $P$ to take a value, say 2,
where $P=2$ if Veronica rescues Betty in such a way that Betty's arm
is not broken.  Reasonable people might disagree whether such an event
is possible.  First suppose we decide that $P$ can take only values 0
and 1.  Then the possible outcomes are that Betty drowns ($O=0$) or
Betty is saved ($O=1$). In this model, any utility function that makes
the utility of drowning worse than that of being saved
would result in Veronica's rescue not harming Betty.}

\emph{Now suppose that we allow $P=2$. Then we would take $O=1$ to
represent Betty being saved but her arm being broken, and $O=2$ to
represent Betty being saved without her arm being broken.  
In that case, whether Veronica harms Betty depends on the default.  If
we take the default utility to be ${\bf u}(O=2)$ 
then Veronica does
cause Betty harm, while if 
we take 
it to be ${\bf u}(O=0)$, she does not.
Note that the latter choice is quite defensible.  
Given Betty's situation, making it out alive in whatever way possible would presumably be all that matters to her.
} \hfill \wbox
\end{example}
\vspace{-0.2cm}
This example clearly shows that to apply our framework in practice, it is
important to have some guidelines on what count as a reasonable choice,
both in the choice of variables and values and the choice of default
value.  As we mentioned in the introduction, Halpern and
Hitchock \citeyear{HH10} discuss this issue in the context of causal
models; to the best of our knowledge, this issue has not been
discussed in the context of default values. 
 While this issue is
beyond the scope of the current paper, we should make clear that we
would not, in general, expect there to be a unique ``correct'' model.
As we have said repeatedly, reasonable people can disagree about these
choices.  

There is one final issue we would like to address: why
we consider a contrastive definition rather than just giving a
definition in the spirit of the causal-counterfactual account.
Definition~\ref{def:harm2} explicitly invokes a contrastive outcome
$o'$ whose utility is better than that of the actual outcome.
We could have instead just defined harm as the result of causing an
outcome whose utility is worse than the default.

One reason why we did not do so is that the default utility is not
always achievable, and it would be counterintuitive to say that the
agent was harmed if the outcome 
has a utility lower than the default, even though it is the best
possible outcome. For example, there are diseases for which a surgery
can only provide a temporary cure; 
in this case, a successful surgery gives the patient a temporary
relief, and an unsuccessful surgery results in the patient's
death. While the default utility for the patient, 
as for all people, is to be alive and healthy, saying that a successful surgery harmed the patient seems wrong. 
In fact, defining harm as the result of causing an outcome with the
utility worse than the default provides counterintuitive results even
when the default utility is achievable, as 
the following example demonstrates.



\begin{example}[{\bf Pills}]\emph{Consider the following vignette,
again taken from \cite{CJR21} (where it is
presented as a problem for both the causal-counterfactual and contrastive
causal-counterfactual accounts):} 
\emph{\begin{quote}
Barney suffers from a painful
disease. On Monday, he can either take Pill A or not. On Tuesday, he
will have another choice, between taking Pill B or not. Barney
believes that he will be completely cured just in case he takes only
Pill A, and partially cured just in case he takes both
pills. Accordingly, he takes Pill A on Monday and does not take Pill B
on Tuesday \ldots
He is, however, misinformed about the effects of the
pills. Taking only Pill A causes his disease to be merely partially
cured. If he had taken both pills, he would have been completely
cured. Had he not taken Pill A on Monday, on the other hand, nothing
he could have done later would have produced even a partial cure.
\end{quote}
}

\emph{To capture this in our framework,
let $O$ be a three-valued variable that captures Barney's
health: $O=2$ if he is fully cured, $O=1$ if he is partially cured,
and $O=0$ if he is not cured at all. $A$ and $B$ capture whether or not
Barney takes pills A and B respectively. The equation for $O$ is then
such that $O=2$ if $A = B = 1$, $O=1$ if $A =1$ and $B=0$, and $O=0$
otherwise. As Barney considers taking pill B only if he fails to take
pill A, the equation for $B$ is
$B=\lnot A$.
The context is such
that $A=1$; therefore, $B=0$ and $O=1$.} 

\emph{Carlson et al.~claim that taking the pill does not harm Barney;
we agree. 
Yet it easy to see that $A=1$ does cause $O=1$.  Indeed it is a
but-for cause: had Barney not taken the pill, $O$ would have been $0$.
It is easy to see why this is a problem for the 
causal-counterfactual
account: Barney would have been better off if $O=1$ had not obtained;
specifically, he would be better off if $O$ had been 2 (although this
is not the outcome 
that results when 
changing $A$ to $0$
and therefore is not a problem for the counterfactual comparative account).
Carlson et al.~also
view it as a problem for the contrastive causal-counterfactual
account, because in applying it, they compare $O=1$ to the outcome
$O=2$, 
(which, again, is not the outcome that obtains by switching $A$ to $0$),
since they take the closest world to the one where Barney takes just
one pill to be the world where he takes both pills.
Our definition avoids this problem. We do not consider the ``closest''
state of affairs.  Rather, we compare $O=1$ to the outcome $O=0$ caused by
switching to $A=0$. 
$O=0$ has utility worse than that of the outcome obtained from $A=1$,
so there is no harm according to our 
definition, for what we view as 
the ``right'' reasons. 
Assuming that the default utility is ${\bf u}(O=2)$, $A=1$ does cause an outcome whose utility is worse than the default and therefore
 a non-contrastive version of our definition would not have given the
desired result.
} \hfill \wbox
\end{example}

\end{document}